\documentclass{article}

\usepackage{arxiv}

\usepackage[utf8]{inputenc} % allow utf-8 input
\usepackage[T1]{fontenc}    % use 8-bit T1 fonts
\usepackage{hyperref}       % hyperlinks
\usepackage{url}            % simple URL typesetting
\usepackage{booktabs}       % professional-quality tables
\usepackage{amsfonts}       % blackboard math symbols
\usepackage{nicefrac}       % compact symbols for 1/2, etc.
\usepackage{microtype}      % microtypography
\usepackage{lipsum}		% Can be removed after putting your text content
\usepackage{graphicx}
\usepackage{natbib}
\usepackage{doi}

\usepackage{amsmath}
\usepackage[table]{xcolor}

\usepackage{algorithm}
\usepackage{algorithmicx}
\usepackage{algpseudocode}

\usepackage{tabularray}
\UseTblrLibrary{booktabs}
\usepackage{caption}
\usepackage{float}
\usepackage{placeins}

\title{Fairness through Feedback: Addressing Algorithmic Misgendering in Automatic Gender Recognition}

\author{ \href{https://orcid.org/0000-0002-6474-1284}{\includegraphics[scale=0.06]{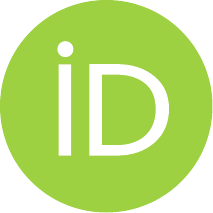}\hspace{1mm}Camilla Quaresmini}\\
	Department of Electronics, Information \\
    and Bioengineering\\
	Politecnico di Milano\\
	Milan, Italy \\
	\texttt{camilla.quaresmini@polimi.it} \\
	\And
	\href{https://orcid.org/0000-0001-9898-4113}{\includegraphics[scale=0.06]{orcid.pdf}\hspace{1mm}Giacomo Zanotti} \\
	Department of Electronics, Information \\
    and Bioengineering\\
	Politecnico di Milano\\
	Milan, Italy \\
	\texttt{giacomo.zanotti@polimi.it} \\
}

\renewcommand{\shorttitle}{\textit{arXiv} Template}

\hypersetup{
pdftitle={A template for the arxiv style},
pdfsubject={q-bio.NC, q-bio.QM},
pdfauthor={David S.~Hippocampus, Elias D.~Striatum},
pdfkeywords={First keyword, Second keyword, More},
}

\begin{document}
\maketitle

\begin{abstract}
Automatic Gender Recognition (AGR) systems are an increasingly widespread application in the Machine Learning (ML) landscape.
While these systems are typically understood as detecting gender, they often classify datapoints based on observable features correlated at best with either male or female sex. In addition to questionable binary assumptions, from an epistemological point of view, this is problematic for two reasons. First, there exists a gap between the categories the system is meant to predict (woman \textit{versus} man) and those onto which their output reasonably maps (female \textit{versus} male). What is more, gender cannot be inferred on the basis of such observable features. This makes AGR tools often unreliable, especially in the case of non-binary and gender non-conforming people. 
We suggest a theoretical and practical rethinking of AGR systems. To begin, distinctions are made between sex, gender, and gender expression. Then, we build upon the observation that, unlike algorithmic misgendering, human-human misgendering is open to the possibility of re-evaluation and correction. We suggest that analogous dynamics should be recreated in AGR, giving users the possibility to correct the system’s output. While implementing such a feedback mechanism could be regarded as diminishing the system’s autonomy, it represents a way to significantly increase fairness levels in AGR. This is consistent with the conceptual change of paradigm that we advocate for AGR systems, which should be understood as tools respecting individuals’ rights and capabilities of self-expression and determination.
\end{abstract}

\keywords{Automatic Gender Recognition \and Gender Bias \and Algorithmic Fairness \and Sexual Identity \and Face Recognition}

\section{Introduction}\label{sec:1}

Automatic Gender Recognition (AGR) systems are an increasingly widespread application in the Machine Learning (ML) landscape \cite{karahan2022age,lin2016human}. As a subfield of Face Recognition (FR), they algorithmically extract gender from facial physical traits to perform classifications \cite{hamidi2018gender,keyes2018}, based on the idea that the face is the most expressive and easily observable part of human bodies \cite{ng2015review,vijayan2016face}. The use of AGR goes from general identity verification to behavior and preference prediction for content recommendations, up to person categorization for surveillance purposes \cite{demirkus2010automated,ng2015review,paul2013human}. Indeed, they are embedded in advertising, security, biometrics, human-computer interaction, and healthcare tools. Depending on the specific application, they can be used to make inferences that directly affect human lives in significant ways. For instance, in healthcare contexts, gender may be used to infer the probability of developing a disease. In hiring processes, instead, candidates' gender may (unfortunately) influence the evaluation.

As already noted in the literature, AGR systems are built on questionable and problematic assumptions which make them often unreliable, especially in the case of interactions with non-binary and gender non-conforming people \cite{scheuerman2019}. What is worse, being misgendered by such automated tools can be painful for individuals whose sexual identity diverges from what is often taken to be the norm (in a strong sense), namely a cisgender and gender-conforming person.

Further analyzing these questions, different limitations of AGR systems can be pointed out. To begin, they treat gender as a static feature. As such, each datapoint is categorical, fitting one gender label which cannot change over time. Additionally, the gender labels are mutually exclusive \cite{quaresmini2024data}. Second, they define gender classification as a binary task in terms of \textit{male} versus \textit{female} categories. To make things worse, companies developing AGR systems typically provide no documentation to clarify if their tools are classifying gender identity or sex \cite{scheuerman2019}. Third, they perceive gender as inferable from physical traits, more precisely by considering the face as a proxy for it. However, far from being directly deducible from images, gender is a complex construct reflecting individuals' intrapsychic aspects, and as such cannot be inferred on the basis of observable facial features \cite{spiel2019}. Due to such simplifications embedded in their architecture, AGR systems can easily lead to mismatches between the theoretical understanding of constructs and their operationalization \cite{wallach2019}. This is particularly problematic when it comes to classifications of non-binary and gender nonconforming people, as shown in \cite{keyes2018,quaresmini2024data,scheuerman2019}.

Despite being limited and questionable, the kind of classification we described is already happening, spreading with commercial systems offering gender classification as a standard feature \cite{scheuerman2020we}. This widespread use of AGR systems pushes us to investigate whether there is a strategy to ensure that the labels assigned during gender classification are as less stereotypical and archetypal as possible, and so the mechanism of gender classification itself. Building upon an inclusive taxonomy of the dimensions constituting sexual identity as well as their interaction, this work suggests a theoretical rethinking of AGR systems as well as practical design strategies for improving their functioning, based on the idea of a feedback mechanism to be embedded in the classification process.

The paper is structured as follows. In Section \ref{sec:2}, we present sexual identity as a composite concept, making distinctions between sex, gender identity and gender expression, with the aim to provide a framework for classification that is more comprehensive than the standard male/female binarism. Having done that, Section \ref{sec:3} presents the functioning of AGR. In Section \ref{sec:4}, we consider the dynamics of interpersonal misgendering and, taking inspiration from the correction mechanisms that can take place within such dynamics, we propose a design solution for improving AGR tools' fairness. Section \ref{sec:5} discusses some implications of our proposal, while Section \ref{sec:6} concludes the work, suggesting future extensions.

\section{Who is afraid of gender?}\label{sec:2}

Long-running critiques of AGR have highlighted the negligence towards important conceptual distinctions. Most notably, technical approaches in AGR often conflate gender with sex, more or less explicitly taking the former to be identical to the latter, or directly determined by it. This section aims to provide the reader with a more in-depth picture on gender, sex, as well as their relations with other components that are relevant for our purpose.
Although conceptually distinct, sex and gender can be understood as different components of a single, multifaceted construct, namely sexual identity \cite{shively1977components}.

According to an unduly oversimplified model of sexual identity, gender is a question of being either male or female. In particular, two features of this taxonomy stand out: its binarism --- it's either male or female, no other options allowed --- and the complete overlap between sex and gender. Most notably, this overlap is evidenced by the male/female dichotomy, typically used to refer to \textit{sex} in all animals, to list \textit{gender} possibilities. However, gender is different from sex, and this difference is typically marked also terminologically. Remaining within a binary perspective, when it comes to gender, we tend to talk about \textit{men} and \textit{women}.

Importantly for our purpose, most AGR systems build upon this oversimplified and inexact model of gender. As we will see in Section 3, most AI tools for gender recognition are designed to output either ``male" or ``female", and existing training and testing datasets only distinguish between two categories of individuals, often using the male \textit{versus} female labeling.

After decades of activism and research in psychological and social sciences, however, we understand sex and gender in a significantly different way. As anticipated, nowadays we conceptualize them as distinct components of what may be called sexual identity (cf. \cite{west1987doing, prentice2002women}).\footnote{Note that in this paper we will not consider all components of sexual identity. Most notably, we will not touch upon sexual orientation, having to do with individuals' sexual and romantic attraction (or lack thereof) towards other people.} Table \ref{table:1} provides a tentative overview of the components of sexual identity that are relevant for our aim, necessarily overlooking aspects and distinctions that would make the discussion overly complex.

\begin{table}[ht]
  \centering
  \caption{Overview of sexual identity components.}
  \label{table:1}
  \begin{tblr}{
    cells      = {c},
    % shade columns 2 and 3 in 20% gray:
    column{2}  = {bg=gray!20},
    column{3}  = {bg=gray!20},
    % your multirow/multicol settings as before:
    cell{1}{2} = {c=2}{},
    cell{1}{4} = {c=2}{},
    cell{2}{3} = {r=2}{},
    cell{3}{4} = {r=2}{},
    cell{3}{5} = {r=2}{},
    vlines,
    hline{1-2,5} = {-}{},
    hline{3}     = {1-2,4-5}{},
    hline{4}     = {1-3}{},
  }
    {\textbf{Sex} (Biological/assigned\\at birth)}
  & {\textbf{Gender} (Not meant to be\\exhaustive)}
  &            & \textbf{Gender expression}
  &                       \\
  Female   & Woman        & Cisgender     & Feminine      & Gender conforming     \\
  Male     & Man          &               & Masculine     & Gender non-conforming \\
  Intersex & Non-binary   & Transgender   &               &
  \end{tblr}
\end{table}

Starting from the left, sex refers to the physiological and biological features of the individual, such as chromosomes, hormones, reproductive organs, and secondary sexual characteristics. Among humans, as it happens for many other species, individuals are taken to be either male or female. And while this may be true for the majority, intersex people exist as well, displaying physical, biological and physiological features that do not perfectly match those typically associated with male and female individuals \cite{preves2003intersex}. \footnote{Interestingly, among intersex people, some are not recognized as such and are categorized as either males or females at the moment of birth on the basis of visible genitalia --- in fact, many individuals find out that they are intersex only in puberty or even later in life. Instead, newborns who are recognized as intersex based on so-called ambiguous genitalia often undergo immediate surgery-mediated sexual reassignation.} In all cases, it is important to distinguish between the biological dimension of sex, that can be conceived of as a spectrum, and what we refer to with the expression ``sex assigned at birth", often reflecting the social pressure to categorize all individuals --- including intersex ones --- as either male or females.

While sex is understood as having a largely biological basis, gender has to do with distinctively social and psychological dimensions (see \cite{morrison2021erasure}). Definitions of gender abound in scientific and public discourse, and providing a univocal characterization is far from an easy task. To unpack this concept, however, we can start from the WHO's definition according to which ``gender refers to the characteristics of women, men, girls and boys that are socially constructed.  This includes norms, behaviors and roles associated with being a woman, man, girl or boy, as well as relationships with each other. As a social construct, gender varies from society to society and can change over time".\footnote{\url{https://www.who.int/health-topics/gender\#tab=tab_1}.} Importantly, gender is not a matter of biological determination, but rather a question of self-identification and perception: if $x$ perceives herself as a woman, then $x$ \textit{is} a woman. 

In this respect, a clarification must be made on the relationship between sex and gender. On the one hand, it is true that the majority of people are \textit{cisgender}. That is, their gender aligns with the sex they were assigned at birth. To be more precise, it aligns with social expectations determined by the sex assigned at birth, according to the equations ``male=man" and ``female=woman". On the other hand, transgender people, who experience an incongruity between their self-perceived gender and the sex they were assigned at birth, are a non-negligible portion of the population.\footnote{Statistics are not particularly reliable, for trans people have often been victims of invisibilization and, also depending on the context, coming out as trans can put the individual at risk.}

Note that the category of trans people does not include only trans women and men --- respectively, people assigned male at birth who identify as women, and people assigned female at birth who identify as men. Here, we take the label ``trans" to apply also to non-binary people, namely those who do not recognize themselves within the man-woman binarism. As many of the terms employed here, ``non-binary" is a label subsuming a number of different experiences, ranging from people perceiving themselves as belonging to more than one gender to people who are fluid in their gender identification and individuals who do not associate any gender with their experience. For this reason, we refer to non-binary individuals to capture a heterogeneous group of people that, despite their differences, share the rejection of gender binarism.

While sex is a matter of (assignation on the basis of) biology, and gender refers to an intrapsychic dimension having to do with the individual's self-perception, the concept of gender expression captures aspects related to a person's behavior and presentation to others. To understand gender expression, we need to start from societal expectations. Depending on the cultural setting, some behaviors and aspects of an individual's appearance are strongly associated with a specific gender. Since we are interested in AGR through face recognition, we focus on faces. In many contemporary Western cultures, for instance, wearing visible makeup is not something most men regularly do. And more importantly, society does not expect them to do it: a man wearing visible makeup is likely to meet dismay or even more or less explicit disapproval. In many contexts, something analogous could arguably be said about a woman wearing her hair short and using no makeup. Here, we cannot afford an exhaustive discussion on gender expression and roles. For the sake of simplicity, in this work, individuals meeting societal expectations in terms of gender expression are referred to as having a \textit{conforming} gender expression, in contrast with people having \textit{non-conforming} gender expression. The point is simply that gender expression is not an infallible indicator of gender: even if social expectations are particularly strong and most people tend to express themselves in a gender-conforming way, a universal assumption of gender conformity cannot be made, and someone's gender expression is not an infallible indicator of their gender.

Sex, gender, and gender expression are thus dimensions which, despite being interrelated, are to be kept distinct and understood as independent from each other. In particular, sex and gender do not overlap, and not all people are cisgender --- or even fitting in gender binarism. In addition to this, gender expression is not always a reliable marker of someone's gender. Needless to say, this has profound implications for AGR, for the inferability of gender from physical features and appearances is not granted. In particular, as we are about to see, this non-inferability is the main source of algorithmic misgendering, which we can define as the misclassification of one's gender output by an AGR system. To be clear, algorithmic misgendering is not a purely technical problem. On the contrary, it is a direct consequence of the fact that we inhabit a strongly cis-normative society, permeated by strong expectations concerning gender expression. 

\section{Automatic Gender Recognition}\label{sec:3}

The discussion on the components of sexual identity we have developed in the previous section could have been much longer. To begin, we have overlooked the component of sexual and romantic orientation, which has little to do with AGR.\footnote{The question of automatic sexual orientation detection is another serious one \cite{DeBlock_Conix_2022}.} More importantly for our purpose, however, we may have conveyed the impression that everybody could (or would want to) be categorized according to the labels we have seen. This, however, is not completely accurate. While it is true that many people can identify with some of the groups and identities we have listed, some people --- typically self-identifying as \textit{queer} --- reject such taxonomies, that are deemed to be too static, partial and constraining. As a result, even if a more fine-grained taxonomy of sexual identity is to be used in the context of AGR, some outputs of AGR tools will never be correct. This is due to the fact that some datapoints are in principle not classifiable. We can therefore introduce a distinction between at least two kinds of datapoints:\footnote{The inspiration for this kind of distinction comes from \cite{fresco2013miscomputation}, although the application is different.}
\begin{itemize}
    \item datapoints that are generally \textit{miscategorized}, but whose classification can be amended;
    \item datapoints that are in principle \textit{non-categorizable}, for some people reject the very idea of a classification based on the labels associated with the components of sexual identity.
\end{itemize}
In this work, we will not further consider non-categorizable datapoints, focusing on those cases which are miscategorized but correctly categorizable in principle. Based on what we have seen so far, these instances of algorithmic misgendering could involve (among others):
\begin{itemize}
    \item trans men or women being misgendered respectively as women and men due to some facial features that are significantly correlated with a specific sex (e.g., the shape of lips, jaw, eyes and cheeks);
    \item people whose gender expression is not conforming, regardless of their being cis- or trans-gender;
    \item non-binary people, who are typically left without a label.
\end{itemize}

\noindent Nowadays, algorithms are systematically used to identify and categorize human identities \cite{beemyn2021sage}. And in general, gender is often perceived as a crucial attribute for performing such classification. We would now like to focus on the main existing approaches to AGR in the context of computer vision, thus excluding methods that analyze features such as voice and walking movements. In particular, we focus on face recognition \cite{bissoon2013gender}, which is typically considered the most prominent ground for such analysis. 

As a classical pattern recognition problem \cite{ng2015review}, the task comprises four distinct steps: 
\begin{enumerate}
    \item object detection;
    \item pre-processing;
    \item feature extraction;
    \item classification.
\end{enumerate}
Given an image, the system (1) detects and crops the region containing a face. Then, during pre-processing, (2) the image is normalized both geometrically (resized and aligned) and with respect to illumination (contrast and brightness). To do this, several methods are proposed in the literature --- see for example \cite{jafri2009survey,oloyede2020review}. External features such as the neck and background are also removed in this phase. At this stage, (3) feature extraction is performed, where representative descriptors of an image are identified. This is followed by a selection of the most discriminative features for detecting gender. Importantly, face recognition technology relies on an appearance-based approach. The most common method to extract features is a geometric-based one which uses facial landmarks (i.e., specific key points on the face) as a guide to detect gender, thus relying on some standards of spatial configuration \cite{koestinger2011annotated,wu2019facial}. Noteworthily, there exist also approaches which analyze the pixel intensity of the entire face image and its texture, such as Local Binary Patterns, Histogram of Oriented Gradients, or Gabor Filters --- for the technical details see \cite{al2020new,hajizadeh2011classification,shan2012learning}.

Crucially for our purpose, the specific facial features that convey gender-related information seem to remain unclear in the literature. While saliency maps have identified the most relevant facial regions for gender classification \cite{berta2023ginn} excluding features such as hairstyle, such features appear to play a crucial role in gender detection \cite{li2012gender,lian2008gender}, even in the systems which explicitly focus on face analysis. The point is that if features such as hairstyle, make-up, clothing, or other non-facial elements are (inadvertently or not) driving classification, models may generalize poorly, encoding cultural biases and social stereotypes.

The final step is the proper (4) classification of the obtained information. Here, the classifier is trained on (and validated with) a dataset. Various classifiers are used in the literature, such as Support Vector Machine, Ada-Boost, Neural Networks and Bayesian Classifier --- for a more technically detailed presentation of such techniques, see \cite{khan2012hybrid,kumar2019gender,kumar2023predictive,levi2015age,liew2016gender,moghaddam2000gender,shih2012gender}.

Now, it is widely acknowledged that AGR systems are biased. As we can see in \cite{buolamwini2018gender}, AGR tools present important accuracy problems affecting dark-skinned women. Indeed, it is shown that famous commercial services offering gender classification --- such as Amazon, Clarifai, IBM, Microsoft --- perform poorly on female individuals with respect to male ones. The gap is not negligible, as the error rate for the most misclassified group (dark-skinned women) is up to 34.7\%, and the one of the most correctly classified group (white-skinned men) is 0.8\%. Furthermore, if we consider a more diverse gender classification going outside the binarism, the results are even worse. In \cite{scheuerman2019}, the authors show how the above-mentioned commercial facial analysis services perform consistently worse on transgender individuals, being completely unable to classify non-binary genders. Table \ref{tab:tpr_performance} is adapted from \cite{scheuerman2019}, highlighting the true positive rate (TPR)\footnote{TPR is the proportion of positive instances that are correctly classified by a system.} for each hashtag for each of the facial analysis services analyzed in the study. Here \#agender, \#genderqueer and \#nonbinary hashtags are omitted.

\begin{table}[h]
    \centering
    \begin{tabular}{lcccccc}
        \toprule
        \textit{Hashtag} & \textbf{Amazon} & \textbf{Clarifai} & \textbf{IBM} & \textbf{Microsoft} & \textbf{All (Avg)} \\
        \midrule
        \#woman & 99.4\% & 95.1\% & 98.6\% & 100.0\% & 98.3\% \\
        \#man & 95.4\% & 98.3\% & 97.4\% & 99.4\% & 97.6\% \\
        \#transwoman & 90.6\% & 77.4\% & 94.3\% & 87.1\% & 87.3\% \\
        \#transman & 61.7\% & 76.0\% & 71.4\% & 72.8\% & 70.5\% \\
        \bottomrule
    \end{tabular}
    \caption{TPR performance per gender hashtags (adapted from \cite{scheuerman2019}).}
    \label{tab:tpr_performance}
\end{table}

Attempts to make AGR technology more inclusive for identities outside binarism can be found in the literature. Beyond many studies that admit and outline the disparities in such systems, also proposing theoretical refinements \cite{scheuerman2018safe,pennisi2024operationalization} and suggesting practical regulations \cite{perilo2024non,scheuerman2020we}, some researchers focus on advancing and implementing technical solutions \cite{mahalingam2013eye,kumar2016robust,vijayan2016face}. For example, in \cite{scheuerman2019}, the problem of label incompleteness is addressed by proposing new labels. The same approach is then extended in \cite{quaresmini2024data} by adding a temporal parameter.
However, even if such solutions enter the algorithmic pipeline at the processing stage by trying to fix some wrong assumptions of AGR systems, none explicitly focuses on the user relationship with the classification system itself, recognizing the fact that gender is a matter of self-identification and determination. In the following, we propose our approach in this direction.

\section{A human-inspired solution to algorithmic misgendering }\label{sec:4}

AGR applications turn out to be problematic (among other things) for their frequent misgendering of trans people --- including non-binary ones --- and gender non-conforming individuals. The systematic misclassification of people belonging to these groups is far from being a mere accuracy problem. On the contrary, it has relevant ethical repercussions in terms of harm to already discriminated populations. As a matter of fact, focusing on the psychological dimension, a growing body of research has shown that misgendering and the related behaviors can have a significantly negative impact on the mental health and well-being of trans individuals \cite{galupo2020every, mclemore2018minority, puckett2023systems}. Consequences can be severe, with outcomes ranging from hyper-vigilance to depression and rise of suicide rates \cite{barr2022posttraumatic, rostosky2022lgbtq, tan2020gender}.\footnote{See also \url{https://williamsinstitute.law.ucla.edu/publications/trans-adults-united-states/}.} Now, misgendering mechanisms are different for binary and non-binary trans individuals. In the former case, as a matter of fact, misgendering consists in the negation of their belonging to the gender they identify themselves with. In the case of non-binary individuals, instead, misgendering involves the erasure of their gender identity and the forcing into binarism. In both cases, however, the point remains that individuals are not respected and recognized in their gender identity.

Also due to the psychological impact that misgendering has on trans and gender non-conforming people, there is little doubt that intentional misgendering is morally reprehensible \cite{kapusta2016misgendering}. In particular, we can think of deliberate use of wrong pronouns or deadnaming --- that is, using the name a trans person received at birth but rejected later as part of their gender affirmation process. These behaviors, in addition to resulting in psychological harm for the individual, qualify as deliberate attacks on someone's well-being and a disrespectful rejection of their capacity for self-affirmation. Now, algorithmic misgendering can hardly be understood as intentional and therefore directly morally reprehensible, if only for the simple reason that AGR systems have no intentions and do not qualify as moral agents. Also, we do not consider the possibility that AGR tools are purposely designed to be discriminatory against trans and gender non-conforming individuals. Still, we maintain that designers, developers and deployers of AGR at least have the \textit{active} responsibility \cite{bovens1998quest} to prevent forms of algorithmic misgendering.

\subsection{Correcting misgendering}

Our discussion will now focus on this question: how can AGR tools be improved? Before trying to answer this question, it is important to make our standpoint clear. On the one hand, we are not arguing that such systems should be employed, nor we wish to somehow defend their legitimacy. On the other hand, we do not aim to propose a straightforward argument to the effect that AGR itself should not be a thing \cite{keyes2018}.\footnote{See also \url{https://campaigns.allout.org/ban-AGSR}.} Our starting point is that, in spite of ethical discussions, the current situation is that AGR tools are developed and widely deployed in a way that is largely dictated by economic interests and rules of the market. Completely banning them is presumably a way to protect people who are already discriminated and invisibilized for their sexual identity. However, the feasibility of this approach is not straightforward. Here, we adopt a more modest approach by considering alternative --- and maybe more easily implementable --- approaches, trying to make AGR more inclusive and ethically acceptable.

As a heuristic for identifying strategies that could mitigate the risk and impact of algorithmic misgendering, we decided to start from interpersonal interactions. As a matter of fact, misgendering is not an issue only for human-machine interaction. Quite the contrary, indeed. In our everyday life, we systematically attribute gender to other people (e.g., in pronoun choice). However, our gender ascriptions tend to be deeply binary- and cis- normative, and strongly dependent on rigid expectations in terms of gender expression. Just to give an idea, \cite{jacobsen2024misgendering} administered a questionnaire on misgendering to trans and non-binary people, revealing that 59\% of non-binary participants were misgendered daily, 30\% weekly or monthly, and 11\% yearly or less. Trans individuals who identified with a binary gender had lower rates. The results are nonetheless relevant, for they were misgendered daily in 25\% of cases, weekly in 43\% and yearly or less in 31\%. Without focusing on the specific percentages, the point is that misgendering is unfortunately very common in everyday interpersonal interactions.

Even assuming that all the reported instances of misgendering were unintentional --- an assumption that, unfortunately, seems hazardous --- the problem is significant. If on the one hand unintentional misgendering does not have the same moral implications of deliberate one, the potential harm to the misgendered individual is still there. This is also the reason why, when we don't know someone's gender identity, the best strategy consists in asking the person in question what their pronouns are. Once the misgendering took place, however, there is little choice beyond trying to reduce the impact of the wrong attribution. This attempt typically takes the form of the correction of the wrong gender attribution. 

Again, this does not cancel the initial misgendering, and there are arguably cases in which apologies and future attention to pronouns cannot make up for the fact that, at the beginning, there was a wrong gender attribution. For example, we can think about the case of a trans man whose aim is \textit{passing} --- that is, being perceived --- as a man, for whom the initial misgendering is a reminder of the fact that people may not perceive him in the way he would want them to. That said, by apologizing for the wrong attribution and inquiring about pronouns, at least we make sure that the person we interact with feels validated and respected in their gender identity, regardless of the sex assigned at birth, their sexual characteristics or their gender expression.

Needless to say, the contexts of deployment of AGR are significantly different from the interpersonal ones in which such corrections can take place. Most notably, we deal with typically non-dialogic interactions between human agents and artificial ones, which cannot feel sorry or apologize in the strict sense --- again, an AI system is not a moral agent. Still, we hold that a reflection on the dynamics of misgendering in interpersonal relations can be quite useful when it comes to designing more inclusive and fair tools for AGR. In particular, we are interested in exploring whether and how a correction procedure could be integrated into AGR tools.

In the next section, we will present a possible strategy to implement a feedback mechanism in AGR tools that is inspired by misgendering correction in interpersonal relations. Again, we do not take a stance with respect to the very legitimacy of such technologies. Rather, we start by acknowledging that AGR systems exist and are increasingly widespread, embedded in a growing number of widely used platforms and tools. On this ground, our aim is to rethink AGR in a way that could make it more inclusive and fair.

\subsection{Fairness through Feedback}

Building on our discussion about the mechanisms underlying misgendering and the potential avenues for correction in interpersonal settings, we now propose the \textit{Fairness through Feedback} (FtF) algorithm for gender recognition, which is based on the interaction between the classification system and the individual that is going to be classified. The focal point is that the suggested algorithm is designed in a way that allows the user to confirm or correct a predicted gender label.

\begin{figure}
    \centering
    \includegraphics[width=0.5\linewidth]{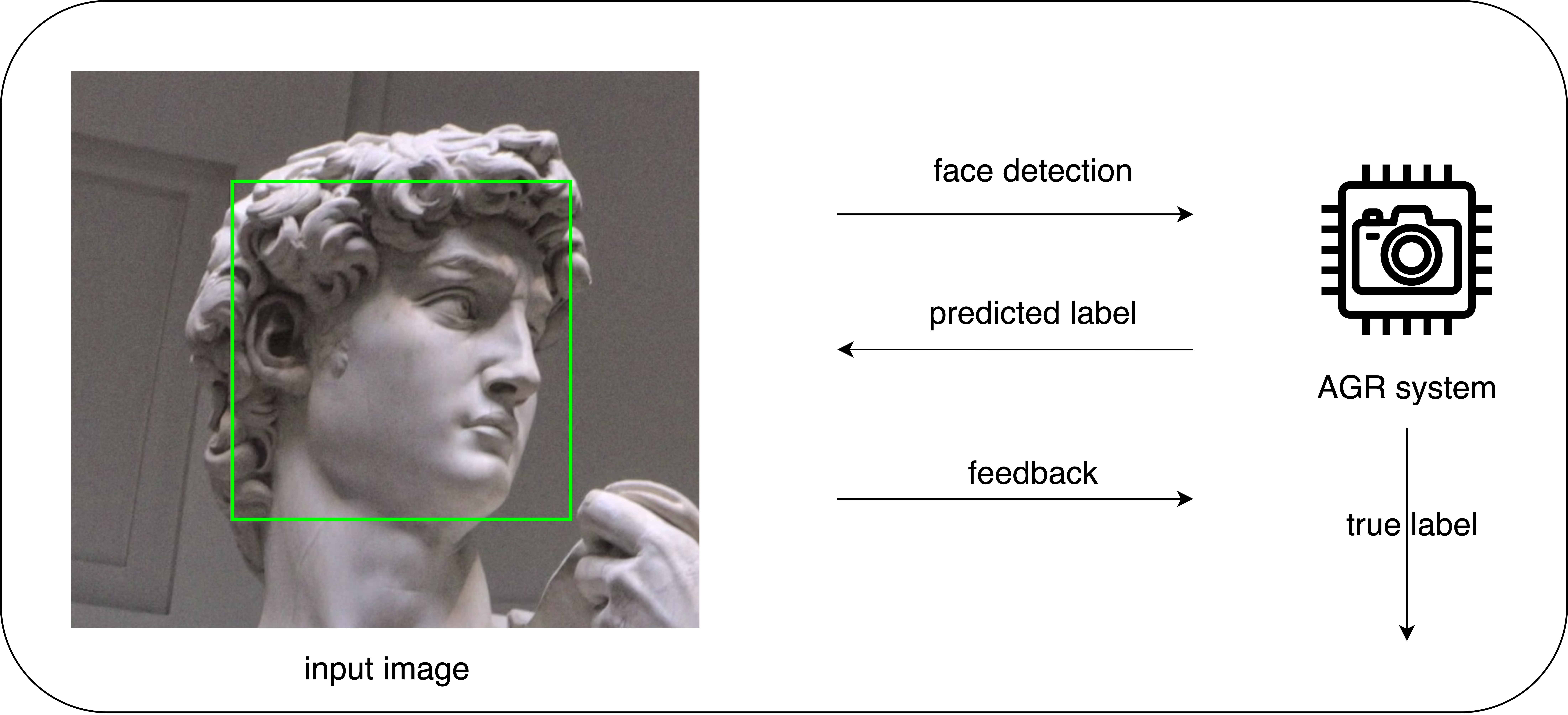}
    \caption{A diagram showing the proposed feedback mechanism.\protect\footnotemark}
    \label{pipeline}
\end{figure}

The system consists of a gender classification system that uses a convolutional neural network (CNN) to classify gender based on facial image input, using MobileNetV2 as a base model (which is pre-trained on ImageNet). It receives as input an image depicting the user's face and works as follows. First, once the input image is analyzed, the system displays the predicted label to provide the user with an initial classification. Then, it asks the user for feedback, by prompting them to either confirm the prediction by leaving the input block blank, or correct the label if the initial prediction was incorrect. If the user does not provide any input within a time frame $t_1$, the predicted label is automatically confirmed. At this stage, the algorithm validates the user's input by checking whether it matches one of the valid labels, i.e., \textit{man}, \textit{woman}, \textit{non-binary}. Importantly, this label set can be easily extended as needed by implementing a larger partition of labels. The last step consists in returning the true label as a corrected output.\footnote{The possibility of users purposely providing an incorrect label will be considered in Section \ref{sec5.1}} This can match or not the originally predicted label. In this way, we ensure that, if the initial prediction is incorrect, the user can provide the correct label instead of being definitively misgendered by the system. An intuitive diagram of the proposed mechanism is depicted in Figure \ref{pipeline}.\footnote{A preliminary prototype has been developed for internal evaluation and is not yet publicly available. We plan to release a complete version in the future.}

\footnotetext{This image is a provisional placeholder provided for illustrative purposes; the final version will include a photograph of one of the paper’s authors.}

\begin{algorithm}
\caption{Get User Feedback}\label{algorithm}
\begin{algorithmic}[1]
\Function{user\_feedback}{predicted\_label, $t_1$}
    \State \textbf{Display predicted label:}
    \State \texttt{print(f"Predicted label: \{predicted\_label\}")} 
    \State \textbf{Set time constraint:} $t_1 = 5$ seconds
    \State \textbf{Get user feedback within } $t_1$:
    \State \texttt{feedback = get\_input\_with\_timeout("If this is not the correct label, enter the correct label (man/woman/non-binary). Leave blank if correct: ", $t_1$)}
    \If{\texttt{feedback} is \textbf{None} \textbf{or timeout occurs}}
        \State \Return \texttt{predicted\_label} \Comment{Automatic confirmation}
    \ElsIf{\texttt{feedback in ['man', 'woman', 'non-binary']}}
        \State \Return \texttt{feedback}
    \Else
        \State \Return \texttt{predicted\_label}
    \EndIf
\EndFunction
\end{algorithmic}
\end{algorithm}

\noindent Some additional remarks are necessary to understand the proposed algorithm's functioning. For the training procedure, as the dataset we use is labelled in terms of male and female datapoints, when datapoints are labelled as \textit{male} or \textit{female}, we consider them respectively to be \textit{man} and \textit{woman}. The use of a binary database is due to two main reasons. First, even if there exist in the literature some attempts in constructing non-binary datasets --- for example \cite{mahalingam2013eye}\footnote{The HRT Transgender Database is no longer available. See \url{https://www.aiaaic.org/aiaaic-repository/ai-algorithmic-and-automation-incidents/hrt-transgender-dataset} for documentation.}, \cite{wu2020gender} and \cite{scheuerman2019} --- they are not publicly available. This is due to privacy issues, as this data are typically scraped by social networks without users consent \cite{keyes2022feeling}, which however undermine their role in making AGR systems more fair. Indeed, in the mentioned cases, datasets are often created ad hoc for the specific study and are exclusively used within that research. Second, we recognize the possibility of creating a new database to include visual images of non-binary individuals by generating synthetic data. However, this paves the way for new potential harm, like reinforcing stereotypical inferences and prejudice related to the visual aspect of such identities \cite{ungless2023stereotypes}.\footnote{See for example the case proposed in \url{https://www.wired.com/story/artificial-intelligence-lgbtq-representation-openai-sora/}, showing how generative tools often provide generated images of non-binary individuals with features like purple hair.} This will be the subject of future investigation.\footnote{Also, conventional data practices have been shown to be inadequate in capturing identities out of gender binarism. However, the present work does not engage directly with this issue. If interested, see \cite{guyan2022queer,ruberg2020data}.}

\subsection{About AGR accuracy and utility}
Importantly, our proposal may give rise to questions in terms of reusability of users' photos and the associated feedback. By default, these are not made available. However, users are given the option to voluntarily choose whether they wish to contribute to the algorithm's training --- again, there is no need to proactively opt out as the consent is not the default option. Here, the purpose of data collection is merely to enhance the accuracy of the system.\footnote{The data collected (i.e., photos and feedback) will only be used for the purpose of enhancing the accuracy of the system. Data will not be utilized for any other purposes, nor will they be shared with third parties without users' consent.} Indeed, since we use a binary dataset for the algorithm's training (and testing), the initial accuracy for the \textit{non-binary} class is zero. However, admitting that an increasing number of non-binary individuals are classified via feedback, the class accuracy is expected to improve. Accordingly, we propose a controlled update, wherein the system's accuracy is periodically assessed at fixed and reasonable time intervals (e.g., annually) to evaluate its performance at each stage. Once an acceptable accuracy level on the class is achieved, then the system is updated.

In light of this, we can draw some theoretical results regarding the general utility of AGR systems. First, from our considerations, it seems to be inversely proportional to time. Indeed, as time passes, the utility of AGR decreases. This is basically due to the potentially increasing incompleteness of the system's label set, as the established categories become out of date over time, thus leading to outdated labels, and new ones have to be constantly added \cite{quaresmini2024data}. This is especially true for the case of binary labels. Conversely, the accuracy of the system increases over time. Indeed, as non-binary datapoints are classified, accuracy increases accordingly.

To put things in slightly more technical terms, the utility of AGR systems over time is proportional to the ratio between their accuracy $A(t)$ and the incompleteness of their label set $L(t)$, such that $U_{\text{AGR}}(t) \propto \frac{A(t)}{L(t)}$ where $A(t)$ is a (monotonically) increasing function representing the accuracy growing as new categories are integrated, and $L(t)$ is an increasing function capturing the degree of incompleteness of the label set increasing over time as gender categories potentially evolve. In other words, the utility of AGR systems depends on their ability to improve accuracy while mitigating the increasing incompleteness of the label set. As a matter of fact, the utility of the system is initially low because accuracy is low and label incompleteness is high. Over time, the accuracy increases. However, if label incompleteness grows at a faster rate, the utility may still decline. The key to maintaining (or improving) utility is to ensure that the accuracy grows at a sufficient rate relative to the label incompleteness.

While this theoretical relationship provides a promising framework for understanding the dynamic trade-offs in AGR systems, its validity remains to be tested empirically. Future research should conduct longitudinal studies measuring $A(t)$, $L(t)$, and $U_{\text{AGR}}(t)$ across different datasets and temporal spans to confirm or refine the proposed model. Additionally, experimental validation should assess whether the rate of accuracy improvement sufficiently compensates for the increasing label incompleteness to sustain or enhance AGR systems' utility over time. Here, our aim was more modest: providing the conceptual and methodological foundations for a different way of implementing AGR, namely the FtF mechanism. As far as we can see, it is crucial regardless of whether users' feedback could fruitfully be employed to put together more representative datasets.

\section{Discussion}\label{sec:5}

Having presented the FtF mechanism and its rationale, we now turn to consider potential objections and open questions. This will allow us to clarify our positions and may provide inputs for further research.

\subsection{A question of context}\label{sec5.1}
In the FtF mechanism, users' feedback ultimately overrides the algorithmic output. Just like the misgendered person's correction in human-human misgendering should be, it is final and indisputable. Again, the \textit{desideratum} behind our proposal is the possibility for all users to feel recognized and respected in their identity, over which they should always have the last word. This may prompt an objection: could one illicitly take advantage of this mechanism to modify the output of an AGR tool to their advantage? For instance, imagine a situation in which being categorized as \textit{woman} gives the user a certain benefit --- e.g., access to reserved parking spots, closer to the entrance, in an underground parking. Could a man misuse the feedback mechanism to be eventually categorized as a woman and thereby get access to the reserved parking spots?

To address this kind of concern, let us make a more general point concerning the contexts in which we take the use of AGR tools to be legitimate. To put some flesh on the bones, let us consider the real-world case of Giggle. This woman-only online platform made the news in December 2022, when the Australian trans woman Roxanne Tickle sued Giggle and its CEO, Sall Grover, after being banned from the app. According to what is reported by several sources that have written on the matter, Tickle had initially been identified as a woman by the AGR tool embedded in Giggle. A few months later, however, she was categorized as a man by a human following a manual review of the updated picture and her profile was blocked, preventing her from further accessing the platform.

Let us leave aside that, in this case, the autonomous recognition system correctly recognized her as a woman, unlike the human operator --- according to many sources, Grover herself --- who blocked her account. The point is: should we allow such uses of AGR tools? Again, high levels of inaccuracy characterize current AGR systems when it comes to trans and gender non-conforming people. With the way things are, is it acceptable that such systems are deployed in contexts in which their output may be the basis for allowing someone to use a service or receive an advantage, while preventing others from doing the same?  Note that we are not questioning the legitimacy of a women-only platform, but rather the acceptability of AGR tools as a way to verify whether someone can access such a platform. Analogous situations are easily conceivable. For instance, we can think of an AGR tool embedded in vending machines that offers women discounts on bus tickets on March 8, in a way similar to Berlin's \textit{Frauenticket}. If the tool misclassified a trans woman as a man, she would end up being prevented from purchasing the discounted ticket in a way we would not hesitate to define as discriminatory.

Given these premises, our assumption is that it should be possible to employ AGR tools only in those cases in which their output does not serve as basis for allowing to or preventing from using a certain service, or receiving a certain advantage.\footnote{For instance, we could think about an online clothing store deploying an automatic gender categorization and thereby immediately directing customers to the associated gendered section of the website. Admittedly, this could give rise to gender discrimination --- e.g., if a trans woman is categorized as a man and automatically directed to the men's section. This risk, however, is averted by the proposed FtF mechanism, for this customer could correct the algorithmic classification \textit{before} it is employed by the website.} Accordingly, the cases we have discussed --- the parking spots, Giggle and the vending machine --- would not be acceptable. While this could be perceived as a significant limitation of AGR's scope, we deem it necessary for maintaining their ethical acceptability. Trans and gender non-conforming people are already systematically discriminated and prevented from accessing services in ordinary contexts, and we should make sure that inaccurate AGR tools do not further aggravate this situation.

At this stage, we can go back to the initial override objection: could one use the FtF mechanism to purposely manipulate the AGR output to their advantage? In the last example we have just made, this scenario could be detailed as the one in which a man ``corrects" the output of the recognition tool, specifying that he is a woman to get the discounted ticket. These situations are prevented by our assumption that AGR tools' output should not be used to allow someone in or leave someone out of a certain service or benefit. Since there would be no perk or advantage following the tool's classification, there would be little point in trying to manipulate it. 

Taking stock on this point, not only limiting the scope of AGR tools mitigates the problem of discrimination following misclassification, but it also makes the override objection against the FtF mechanism largely innocuous.

\subsection{A clash of ideals}
The feedback loop we proposed is meant to increase the fairness of the AGR tools by allowing users to correct misclassifications, giving them the possibility to affirm their gender. However, an objection may come spontaneous: if we introduce such a mechanism in AGR, to what extent can we say that the system actually recognizes gender in a way that can meaningfully be described as \textit{automatic}? Or again, what is the point of having a technology that can autonomously recognize gender if we then posit the necessity of feedback from the user? Could we not simply ask users to provide information about their gender identity?

In fact, the introduction of the feedback mechanism results in a significantly \textit{intrusive system} \cite{ng2012vision} that relies on the cooperation of the user. True, the system still needs to be able to perform AGR, and we aim to achieve high levels of accuracy regardless of the fact that the user can correct the output in a second moment. However, for the output of the system to be finalized and potentially employed for following computations, it needs to be validated by the user. Leaving aside that this procedure could be realistically implemented only on mobile devices --- most notably, smartphones ---  in apps like social-media and e-commerce platforms, this seems to represent a significant reduction of the system's autonomy. This reduction in autonomy could also be interpreted as a waste of resources: Why devote time and money to design a system that could, in principle, work with no human intervention,  if a human input will be nonetheless required? In addition to this, the human intervention step has an impact on the system's efficiency, especially in terms of required time. As a matter of fact, assuming that $t$ is the amount of time a given AGR tool takes to perform a classification on a certain datapoint, the same tool with the FtF module will take at least $t + t_1$, where $t_1$ is the time the user takes to provide the feedback.

To justify our approach on practical grounds, we may insist on the possibility it gives to use (with consent) data from users' feedback to assemble more inclusive datasets. That said, in front of the above-mentioned objections, we largely bite the bullet. That is, we acknowledge that implementing the FtF mechanism will most likely result in a diminished efficiency of AGR tools, and even more importantly that it could represent a reduction of their autonomy. However, unlike our potential objectors, we interpret all of this as a reasonable implication of our proposal, rather than the basis for rejecting it. The reason behind this fundamental disagreement can largely be reduced to a clash of ideals. On the one hand, standard AGR tools not requiring any kind of feedback from their users score greatly in terms of efficiency and performance resulting from their autonomy. This, however, comes at the expense of fairness through hardly correctable algorithmic misgendering. Implementing FtF mechanisms, on the other hand, promotes fairness and mitigates the problem of algorithmic misgendering by giving users the possibility to be associated with the correct label and feel recognized in their identity. Diminished autonomy and reduced efficiency are the cost of this fairness improvement.

Providing a general argument to the effect that fairness --- and not efficiency --- should be our guiding ideal in the design and deployment of AGR tools goes beyond the scope of this paper. By proposing the FtF mechanism, we clearly positioned ourselves on the fairness side. Needless to say, someone prioritizing efficacy would probably dismiss our solution, contending that we should seek strategies to improve AGR tools' fairness without thereby impacting on their performance. How this can be done without our human-inspired FtF mechanism, however, is to be seen. As a matter of fact, the fundamental point remains that external features of individual are not a necessarily reliable guide to inferring gender, and current AGR tools can hardly detect gender on other grounds.

Taking stock on this point, potential criticisms insisting on FtF mechanisms reducing AGR tools' autonomy and efficiency allow us to highlight the trade-off between fairness and performance in AGR, as well as the underlying clash of ideals. We maintain that, when it comes to AGR, algorithmic misgendering's potential for psychological harm and discrimination against vulnerable populations should push us to prioritize fairness and respect for the individual, even if they are at odds with other technical ideals and goals.

\section{Conclusions}\label{sec:6}

In the present work, we have analyzed Automatic Gender Recognition systems through the lens of algorithmic misgendering. Despite their widespread use in commercial applications, the classifications these technologies perform may easily turn out to be particularly harmful to certain identities. Furthermore, unlike human-human misgendering, current AGR tools do not allow the correction of their classifications in case of algorithmic misgendering. To address this issue, we propose an algorithm that prioritizes user interaction for label correction, enabling dynamic updates of misclassified datapoints. This approach challenges the traditional use of AGR systems as tools of oppression \cite{scheuerman2020we}, positioning them as instruments for self-definition.

Future work will explore the implications of the correction process we propose through the framework of epistemic injustice \cite{Fricker2007-FRIEIP}, recognizing that placing the burden of correction on marginalized individuals may itself be unjust. Additionally, further research will focus on refining the algorithm, particularly in determining an appropriate time frame for updating label partitions, to be established through policy decisions.

%\bibliography{sn-bibliography}

\bibliographystyle{unsrtnat}
\bibliography{references}

\end{document}